\documentclass[final,5p,times,twocolumn]{elsarticle}

\usepackage{stfloats}

\usepackage{graphicx}
\usepackage{subfigure}
\usepackage{multirow}
\usepackage{wrapfig}

\usepackage{tabularx}
\usepackage{booktabs}
\usepackage{makecell}

\usepackage[utf8]{inputenc}
\usepackage{amsmath}
\usepackage{amsthm}
\usepackage{amsfonts}
\usepackage{epsfig}
\usepackage{psfrag}
\usepackage{pstricks}
\usepackage{algorithm}

\usepackage{algorithmicx}
\usepackage{algpseudocode}  
\floatname{algorithm}{Algorithm}

\graphicspath{{./Figures/}}

\usepackage{amssymb}
\usepackage{bm}

\biboptions{comma,sort&compress}

\usepackage{hyperref}
\hypersetup{
            colorlinks=true,
            linkcolor=blue,
            anchorcolor=blue,
            citecolor=blue
            }

\usepackage{enumitem}
\setlist{nosep,topsep=-\parskip}

\usepackage{color}
\usepackage{ulem}

\journal{Under Review}

\begin{document}

\begin{frontmatter}

\title{AirfoilGen: A valid-by-construction and performance-aware latent diffusion model for airfoil generation}

\author[]{Zhijie Yang}
\author[]{Min Tang}
\author[]{Peng Du}
\author[]{Qiang Zou\corref{cor}}\ead{qiangzou@cad.zju.edu.cn}

\cortext[cor]{Corresponding author.}
\address{State Key Laboratory of CAD$\&$CG, Zhejiang University, Hangzhou, 310027, China}

\begin{abstract}
Airfoil shape design is a fundamental task in aerospace engineering, with a direct impact on flight stability and fuel consumption. Deep learning has recently emerged as a promising tool for this task, but existing deep generative approaches remain limited in both geometric validity and physical controllability. They offer little control over the generated shapes, yielding invalid geometries, and they typically do not condition effectively on aerodynamic performance. To address these issues, this paper proposes AirfoilGen, a valid-by-construction and performance-aware latent diffusion model for airfoil. It first introduces a novel airfoil representation scheme, the circle sweeping representation, to constrain the generative process so that output shapes respect essential airfoil characteristics. It then enables explicit control over aerodynamic performance (e.g., lift and drag coefficients) by operating in a learned latent space: a transformer model encodes airfoil shapes into vector embeddings, and a conditional diffusion model denoises Gaussian noise into these latent embeddings while incorporating target aerodynamic performance. In addition, this paper presents a new dataset of over 200,000 airfoils, which is substantially larger than the widely used UIUC airfoil dataset (1,650 airfoils) and more suitable for training modern deep generative models. Experiments demonstrate that AirfoilGen enables airfoil generation with far greater geometric validity and aerodynamic performance controllability than previously achievable, with an average performance-conditioning accuracy of 98.41\%. 

\end{abstract}


\end{frontmatter}


\section{Introduction}
\label{sec:intro}

Aerospace engineering is central to industries such as transportation, defense, and space exploration. A critical task in this field is airfoil design, which concerns modeling the cross-section of a wing or blade to generate lift forces. Well-designed airfoils can achieve high aerodynamic performance, contributing to both flight stability and fuel consumption~\cite{masters2015review}.

Traditionally, airfoils are designed using parametric methods~\cite{masters2015review}, which define shapes through specific parameters or mathematical functions. While effective, these approaches require substantial expert knowledge to work satisfactorily and are restricted to a narrow, routine design space~\cite{wang2023airfoil}. Recently, artificial intelligence (AI)-based methods have been applied to airfoil design~\cite{brunton2021data}. These methods can explore a significantly larger design space and generate novel airfoil shapes beyond the reach of traditional parametric approaches.

Despite their demonstrated ability to produce novel airfoils, current AI-based methods typically work in a random generative manner, and little attention has been paid to controlling the generation process to obtain airfoils with desired geometries and physics. As a result, invalid airfoil shapes could be generated. 
For example, the recent BézierGAN method~\cite{chen2020airfoil} and diffusion-based method~\cite{wagenaar2024generative} generated airfoils that bear little resemblance to real designs. In practice, to generate usable lift forces, airfoil shapes must satisfy certain geometric characteristics, such as non-self-intersecting and streamlined (see Sec.~\ref{sec:method} for further discussion of these characteristics). Even if a valid airfoil shape is successfully generated, its physical performance is often left uncontrolled. When such control is absent or ineffective, the generated airfoils could fail in application even if their shapes appear plausible.

\begin{figure}[t]
  \centering
  \vspace{15pt}
  \includegraphics[width=0.7\columnwidth]{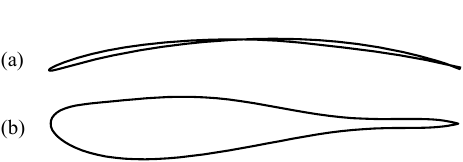} 
  \label{fig:invalid shape}
  \caption{Examples of invalid airfoil shapes generated by BézierGAN~\cite{chen2020airfoil} (a) and the diffusion-based method~\cite{wagenaar2024generative} (b).} 
  \vspace{-10pt}
\end{figure}

To address the above challenges, this paper presents AirfoilGen, a generative model that ensures valid airfoil shapes while enabling performance-aware generation. We first introduce a novel airfoil representation scheme, the circle sweeping representation (CS-Rep), which not only captures the essential characteristics of airfoils but also automatically constrains the generative model's output to respect these characteristics (and therefore stay valid). Unlike commonly used profile-based representations, which are prone to geometric invalidity such as self-intersections, kinks, or distortions, CS-Rep explicitly encodes the airfoil's evolution of thickness and camber through varying circle radii and positions, reducing the risk of invalidity. This structured scheme also makes it easy to impose geometric constraints on the parameters, such as gradual changes, concavity, and monotonicity. As such, airfoil-like geometries are guaranteed by construction.

To enable performance-controlled generation of airfoils, this work first establishes the association between shape and performance in a latent space. An autoencoder is used to transform airfoil shapes into latent vector embeddings, which can be decoded back into CS-Rep format. This latent space provides a common place where geometry and physics can interact, making it possible for AI models to learn complex correlations between shape and aerodynamic performance. A conditional diffusion model~\cite{ho2021classifier} is then trained in this space to guide the latent vector embeddings toward target performance specified by the designer, enabling performance-aware generation. Without this step, airfoils may appear geometrically plausible but fail to meet performance requirements. By combining the structured CS-Rep representation, latent association, and conditional diffusion, AirfoilGen produces geometrically valid and performance-aware airfoils, addressing the two main limitations of existing methods.

Modern generative models achieve strong results with large datasets~\cite{bagazinski2023shipgen, zou2025splinegen}, but the widely used UIUC airfoil dataset~\cite{selig1996uiuc} contains only 1,650 airfoils, which are insufficient for deep learning-based methods. To overcome this limitation, we construct a new airfoil dataset with over 200,000 airfoils, enabling the model to learn diverse airfoil geometries and associated performances.

The main contributions of this paper are:
\begin{itemize}
    \item A novel airfoil representation (CS-Rep) that implicitly encodes essential airfoil shape characteristics and automatically constrains the generative process to ensure valid airfoil shapes by construction.
    \item A latent diffusion model that links geometry and physics, enabling performance-aware airfoil generation.
    \item A large airfoil dataset with over 200,000 models and performance labels, supporting effective training of deep learning models.
\end{itemize}

The remainder of this paper is organized as follows: Sec.~\ref{sec:related work} provides a review of related literature. Sec.~\ref{sec:method} elaborates the proposed AirfoilGen method. Validation of the method using a series of examples and comparisons can be found in Sec.~\ref{sec:result}, followed by conclusions in Sec.~\ref{sec:conclusion}.

\section{Related Work}
\label{sec:related work}

This section summarizes both traditional airfoil design methods from the perspective of CAD~\cite{zou2023variational} and recent generative airfoil design methods from the point of view of AI~\cite{brunton2021data}, as well as the application of generative AI methods to 3D shape design, which is closely related to this work.

\subsection{Traditional Airfoil Design Methods}
\textbf{Parametric methods}. Methods in this category directly parameterize airfoil geometries using predefined shape priors, such as splines~\cite{braibant1984shape}, PARSEC~\cite{sobieczky1999parametric}, and class-shape transformations~\cite{kulfan2006fundamental, he2019improved}. By providing shape parameterizations, they enable effective generation of families of similar airfoils through varying parameters. Nonetheless, their dependence on fixed shape parameterization inherently constrains the design space and makes it challenging to generate novel airfoil shapes.

\textbf{Deformation-based methods.} These approaches generate new airfoil geometries by deforming existing shapes. Early work by Pickett et al.~\cite{pickett1973automated} and Hicks et al.~\cite{hicks1978wing} expressed deformations as linear combinations of simple basis functions. Although computationally convenient, this way of working imposes strong limitations on the design space and leads to unintuitive shape adjustments. More recent work~\cite{kenway2017buffet, leloudas2018constrained} has adopted free-form deformation techniques, which manipulate control points embedded within a surrounding lattice. While these methods provide greater flexibility in capturing complex geometric variations, the control points are often tuned heuristically, such as through area-preservation constraints~\cite{leloudas2018constrained}. Consequently, the resulting design space remains constrained.

\subsection{AI-Based Airfoil Design Methods}
\textbf{Unconditional airfoil generation.} These methods generate airfoil shapes directly, without conditioning on performance or geometric attributes. Early work represented airfoils using spline-based parameterizations and applied generative models to produce the spline coefficients. BézierGAN~\cite{chen2020airfoil}, for instance, combines Bézier curves with a generative adversarial network (GAN). However, GANs are prone to training instability and mode collapse, limiting both robustness and shape diversity~\cite{srivastava2017veegan}. AirfoilGAN~\cite{wang2023airfoil} mitigates these issues by adopting a VAE–GAN hybrid, which improves diversity but still produces airfoils of inconsistent quality. Recent studies~\cite{wei2024diffairfoil, graves2024airfoil} employ diffusion models to improve fidelity, yet the challenge of guaranteeing geometric validity remains largely unexplored.

\textbf{Conditional airfoil generation.} Conditional generative models aim to produce airfoils that satisfy specified geometric or performance criteria. Early approaches~\cite{yilmaz2020conditional, tan2022airfoil} employed conditional GANs that provided only coarse, binary-level control over performance targets. Although these methods demonstrated the feasibility of performance-aware generation, their simplified conditioning schemes limit practical applicability. Subsequent work, such as CEBGAN~\cite{chen2022inverse} and diffusion-based models~\cite{wei2024diffairfoil, wagenaar2024generative, graves2024airfoil}, introduced more precise conditioning on aerodynamic responses, including lift and drag. Nevertheless, controllability remains limited, and the resulting airfoils can still exhibit invalid or impractical geometries.

\subsection{3D Generative Methods}
Airfoil generation is closely related to general 3D shape generation, where the typical approaches may fall into two categories: discrete and continuous. Discrete methods generate voxels, implicit fields, point clouds, or meshes. They commonly employ GANs, VAEs, and, most recently, diffusion models to produce increasingly detailed shapes, e.g., voxels~\cite{wu2016learning, tatarchenko2017octree, ren2024xcube}, implicit fields~\cite{chen2023single}, point clouds~\cite{valsesia2018learning, mo2019structurenet, vahdat2022lion}, and meshes~\cite{groueix2018papier, siddiqui2024meshgpt, Liu2023MeshDiffusion}.

Continuous methods focus on boundary representation (B-rep) models~\cite{zou2019push}, which provide explicit geometric and topological structure~\cite{zou2025bringing}. Methods in this category either generate sequences of modeling operations that yield the desired B-rep model using autoregressive transformers~\cite{wu2021deepcad} or directly synthesize B-rep primitives, i.e., vertices, edges, and surfaces, through structured generative models such as GNNs, transformers, and diffusion models~\cite{jayaraman2023solidgen,zhang2025diffusion}. These approaches offer valuable foundations for airfoil shape modeling, particularly in terms of shape validity~\cite{zou2024intelligent}.

Collectively, the progress in 3D generation highlights a clear shift toward autoregressive transformers and diffusion models. This trend motivates the use of similar architectures for the airfoil generation task considered in this work. However, directly adopting existing models is insufficient, and further developments are needed to ensure geometric validity and controllable shape generation.

\section{Methods}
\label{sec:method}

This work aims to address two key challenges in airfoil generation: ensuring the validity of generated shapes and achieving the desired aerodynamic performance. To ensure the geometric validity, we introduce a novel circle sweeping representation (CS-Rep) for airfoil geometries in Sec.~\ref{sec:method representation}. Then, an autoencoder architecture is utilized to obtain the neural representation that captures the geometric characteristics of airfoils, as described in Sec.~\ref{sec:method embedding}. Finally, we combine the neural representation with the conditional diffusion model to generate airfoils that satisfy aerodynamic constraints in Sec.~\ref{sec:method generation}.

\subsection{Circle Sweeping Representation of Airfoil Shapes}
\label{sec:method representation}

\begin{figure}[htbp]
    \centering
    \includegraphics[width=0.48\textwidth]{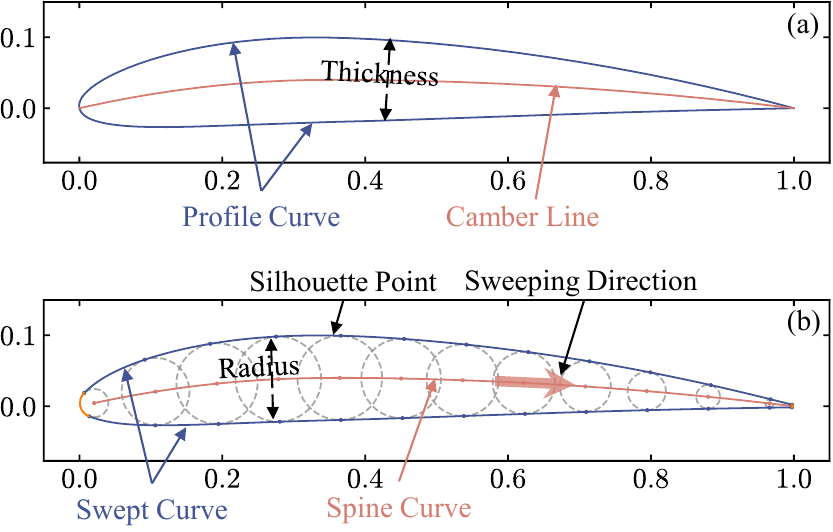}
    \caption{Illustration of CS-Rep: (a) airfoil profiles, and (b) the sweeping process.}
    \label{fig:airfoil shape example}
\end{figure}

An airfoil shape must conform to certain geometric characteristics to produce usable lift. A typical airfoil profile is given in Fig.~\ref{fig:airfoil shape example}(a), and its key characteristics include: 
\begin{itemize}
    \item \textbf{Geometric characteristic \#1:} A single, closed, and simple (non-self-intersecting) profile topology;
    \item \textbf{Geometric characteristic \#2:} a rounded nose region and a sharp, converging tail;
    \item \textbf{Geometric characteristic \#3:} Monotonic progression of the profile geometry along the camber direction to avoid backward-facing steps, folds, or severe concave regions that induce flow separation; and
    \item \textbf{Geometric characteristic \#4:} A streamlined profile with a continuous and smoothly varying thickness distribution along the camber direction, typically exhibiting a maximum thickness in the forward portion and monotonically tapering toward the trailing edge.
\end{itemize}

To capture these characteristics by construction, we introduce a novel airfoil representation scheme called the Circle Sweeping Representation (CS-Rep). Unlike existing approaches that directly use the static airfoil profile (i.e., the contour), this work defines an airfoil as the envelope of a family of circles with varying radii swept along a spine curve, thereby representing the shape through a dynamic construction process. This idea is closely connected to medial-axis-based shape representations, which are well established in geometric modeling, but such structured geometric constructions have not been explored for airfoil generation. By progressively imposing simple yet geometrically motivated constraints on the circle radii and spine geometry, CS-Rep inherently satisfies the above geometric characteristics. As a result, when a generative model operates within the CS-Rep parameter space, the generated airfoil shapes are valid by construction.

\textbf{CS-Rep definition.}
Let $c(t) \in \mathbb{R}^2$ denote a planar spine curve and $r(t) > 0$ a scalar radius function defined along the spine. The airfoil boundary is defined as the envelope of a circle of radius $r(t)$ swept along $c(t)$, as illustrated in Fig.~\ref{fig:airfoil shape example}(b).

The swept curve is obtained from the silhouette points of the moving circle. A point on the airfoil boundary can be expressed as:
\begin{equation}
\label{eq:continuous sweeping}
    p(t) = c(t) + r(t)\, d(t),
\end{equation}
where $d(t)$ is a unit direction determined by the envelope condition:
\begin{equation}
\label{eq:envelope condition}
\frac{d}{dt} \big( \|p(t)-c(t)\|^2 - r(t)^2 \big) = 0.
\end{equation}

Solving Eqs.~\eqref{eq:continuous sweeping} and~\eqref{eq:envelope condition} yields:
\begin{equation}
\label{eq:radius direction}
d(t) = - \frac{\dot{r}(t)\, \dot{c}(t)}{\|\dot{c}(t)\|^2}
\pm \sqrt{1 - \frac{\dot{r}(t)^2}{\|\dot{c}(t)\|^2}} \, e(t),
\end{equation}
where $e(t)$ is the unit normal direction of the spine curve.

\textbf{Discrete CS-Rep for generative AI.}
Since the generative model operates on discrete sequences, we adopt a discretized form of CS-Rep:
\begin{equation}
\label{eq:discrete sweeping}
p_i = c_i + r_i \left(
-\frac{\Delta_c r_i\, \Delta_c c_i}{\|\Delta_c c_i\|^2}
\pm \sqrt{1 - \frac{(\Delta_c r_i)^2}{\|\Delta_c c_i\|^2}}\, e_i
\right),
\end{equation}
where $\Delta_c$ denotes the central difference operator. The spine curve is uniformly sampled along the $x$-axis (in Fig.~\ref{fig:airfoil shape example}(b)) with spacing $\delta_x$.

\textbf{Constraints for geometric characteristic~\#1.}
The sweeping process produces a closed boundary by construction. To avoid self-intersections, the spine curve is constrained to progress monotonically from the leading nose to the tail. Under this constraint, the spine coincides with the medial axis of the airfoil, and this coincidence ensures a single, closed, and simple profile, corresponding to Geometric characteristic~\#1.

\textbf{Constraints for geometric characteristic~\#2.}
A rounded nose region is inherently produced by CS-Rep, as the airfoil is generated by initiating the circle sweeping process at the head. A sharp, converging trailing edge is enforced by imposing appropriate boundary conditions on the radius function $r(t)$: the radius attains its maximum near the leading edge\footnote{The precise location of this maximum is learned and then predicted by the generative model, refer to Sec.~\ref{sec:method shape decoder}.} and decreases smoothly along the camber direction, approaching zero at the trailing edge. Together, these constraints directly ensure Geometric characteristic~\#2.

\textbf{Constraints for geometric characteristic~\#3.}
To prevent backward-facing steps or geometric folds, the $x$-coordinates of the spine points are required to be strictly increasing. This constraint guarantees monotonic progression of the airfoil contour along the camber direction, addressing Geometric characteristic~\#3.

\textbf{Constraints for geometric characteristic~\#4.}
To ensure that the airfoil exhibits a maximum thickness in the forward portion of the camber line and tapers monotonically toward the trailing edge, unimodal constraints are imposed on the radius sequence $r_i$. Specifically, the radius is constrained to increase from the leading edge to a single maximum at a learnable location $pos_r$ in the forward region, and then decrease monotonically toward the trailing edge. This guarantees a physically realistic thickness distribution consistent with conventional airfoil geometries.

\begin{figure}[htbp]
    \centering
    \includegraphics[width=0.4\textwidth]{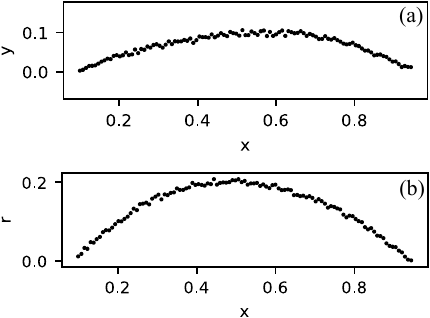}
    \caption{Example of jagged CS-Rep sequences by direct generation: (a) spine sequence, and (b) radius sequence.}
    \label{fig:jagged shape}
\end{figure}

Then we have to avoid abrupt changes in thickness. To ensure smooth airfoil geometry, both the spine sequence and the radius sequence are required to vary smoothly along the camber direction. However, directly generating these sequences may result in jagged shapes, as illustrated in Fig.~\ref{fig:jagged shape}. To address this issue, smoothness is enforced by bounding the increments between neighboring elements in the spine and radius sequences, ensuring gradual geometric variation.

Specifically, the spine curve and radius are represented using difference-based formulations: second-order differences are used for the spine to control curvature variation, while first-order differences are sufficient for the radius. The sequences are defined as:
\begin{equation}
\begin{aligned}
x_i &= x_1 + (i-1)\,\delta_x, \\
y_i &= y_1 + \sum_{j=1}^{i-1} \Delta y_j, \\
\Delta y_i &= \Delta y_1 + \sum_{j=1}^{i-1} \Delta^2 y_j, \\
r_i &= r_1 + \sum_{j=1}^{i-1} \Delta r_j,
\end{aligned}
\end{equation}
where $x_i$ and $y_i$ denote the coordinates of the spine points, and $\Delta$ represents the forward difference operator. Smoothness is enforced by restricting the allowable magnitudes of the second-order spine increments and first-order radius increments:
\begin{equation}
|\Delta^2 y_j| < thres_y, \qquad |\Delta r_j| < thres_r.
\end{equation}

To further prevent geometric distortions such as large fluctuations, local waviness, or multiple unintended extrema, additional unimodal and convexity constraints are imposed on the spine and radius sequences. These constraints eliminate small-scale oscillations that may persist even under smoothness bounds.

Specifically, unimodal behavior is enforced by rewriting the spine and radius sequences as piecewise functions with a single extremum:
\begin{equation}
\label{eq:piecewise reformulation}
\begin{aligned}
\Delta y_i &= 
\begin{cases}
\Delta y_1 + u_i\,(\Delta y_{pos_p} - \Delta y_1), & i \leq pos_p, \\
\Delta y_{pos_p} + u_i\,(\Delta y_{n-1} - \Delta y_{pos_p}), & \text{otherwise},
\end{cases} \\[1em]
r_i &= 
\begin{cases}
r_1 + v_i\,(r_{pos_r} - r_1), & i \leq pos_r, \\
r_{pos_r} + v_i\,(r_n - r_{pos_r}), & \text{otherwise},
\end{cases}
\end{aligned}
\end{equation}
where $pos_p$ and $pos_r$ denote the locations of the extrema of the spine and radius sequences, respectively, and $u_i, v_i \in [0,1]$ are interpolation coefficients.

To enforce unimodality and convexity, the coefficients in each piecewise segment are required to be non-decreasing. A non-decreasing sequence $\{a_i\}_{i=1}^m$ in the range $[0,1]$ can be constructed using a cumulative product formulation:
\begin{equation}
\label{eq:cumulative trick}
a_i = 1 - \prod_{j=1}^i \tilde{a}_j,
\end{equation}
where $\tilde{a}_j \in [0,1]$. This formulation guarantees monotonic progression while remaining compatible with autoregressive generation.

Together, these constraints ensure a smooth, unimodal, and distortion-free thickness distribution, fully satisfying Geometric characteristic~\#4.

\begin{figure*}[htbp]
    \centering
    \includegraphics[width=0.6\textwidth]{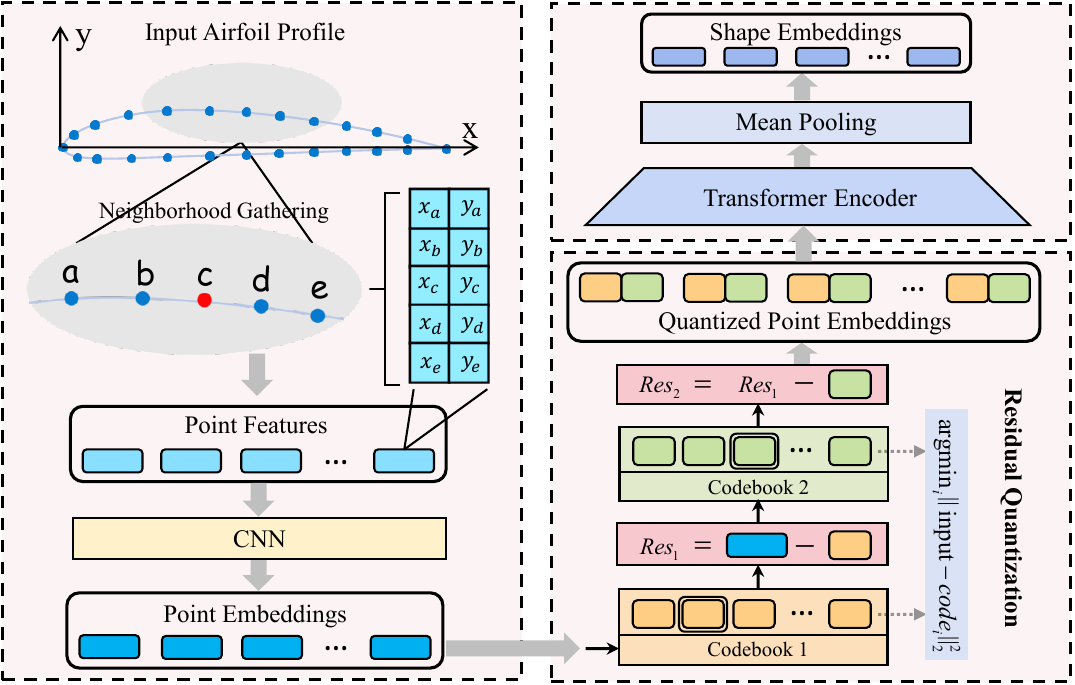}    
    \caption{The network for encoding airfoil profiles into shape embeddings.}
\label{fig:net encoder}
\end{figure*}

\subsection{Airfoil Shape Embedding} 
\label{sec:method embedding}

To provide a more structured representation for generative modeling, an autoencoder is utilized to encode airfoils into latent vector embeddings, denoted as airfoil shape embeddings. Specifically, the autoencoder consists of a residual vector quantization module, a shape encoder, and a shape decoder. The quantization module encodes airfoil profiles into low-dimensional embeddings while preserving precision, facilitating the learning of a compact representation. Together with the shape encoder, this process yields airfoil shape embeddings (Fig.~\ref{fig:net encoder}). Given the airfoil shape embeddings, the shape decoder reconstructs airfoils in the form of CS-Rep (Fig.~\ref{fig:net decoder}).

\subsubsection{Residual Vector Quantization Module}
\label{sec:method rlfq}

The residual vector quantization module takes as input x- and y-coordinates of airfoil profiles as 2-channel features. To represent local airfoil shapes, neighboring points are gathered by concatenating them together, as follows:
\begin{equation}
   o_i = Concat\big(p_{i-\frac{k-1}{2}}, \dots, p_i, \dots, p_{i+\frac{k-1}{2}}\big),
\end{equation}
where $k$ is the size of neighboring points. Given a residual vector quantizer $RVQ$ and a codebook $\mathcal{C}$ with depth $s$, the point embedding $o_i$ is then mapped to the code dimension and quantized by its nearest code in the codebook hierarchically to obtain the quantized point embedding:
\begin{equation}
    h_i = (h_i^1, h_i^2, ..., h_i^s) = RVQ(o_i;\mathcal{C}).
\end{equation}
where $h_i^j$ denotes a code in the $j$-th layer of the codebook $\mathcal{C}$. $k = 5$ and $s = 2$ are set for all cases in this work.

To train the residual vector quantizer and obtain the quantized point embedding $h_i$ that is meaningful for local airfoil shapes, an encoder-decoder scheme is utilized to reconstruct the original neighboring points $o_i$. For stability of training, the decoder adopts a 1-D ResNet. The whole quantization network is trained with the weighted sum of two losses, the reconstruction loss and the codebook loss: the reconstruction loss is the mean squared error between the ground truths and predicted coordinates of neighboring points, and the codebook loss is the loss for the residual vector quantizer to update the codebook (refer to ~\cite{yu2024language} for more details). Based on our experiments, weights 1.00 for the reconstruction loss and 0.01 for the codebook loss yield the optimal results.

\subsubsection{Shape Encoder}

While point embeddings learn the local airfoil shapes, the shape encoder aims to further encode all the quantized point embeddings of an airfoil into a global shape embedding in the unified latent space. The key challenge lies in capturing the global information within the sequence of quantized point embeddings. To model such long-range dependencies, the transformer is a natural choice, because its self-attention mechanism directly considers correlations between all pairs of input embeddings.

Given the quantized point embedding $h_i$, a linear projection $\phi(\cdot)$ is applied to align it with the feature dimension of the transformer encoder. To capture the ordering of the quantized point embedding sequence,  a sinusoidal positional encoding is then added to the projected embedding to obtain the input embedding sequence of the transformer:
\begin{equation}
    f_i = \phi(h_i) + \mathrm{PE}_i.
\end{equation}
The above embedding sequence then passed to a transformer encoder $\mathbf{E}$ to obtain a new embedding sequence containing global information:
\begin{equation}
    \{g_i\}_{i=1}^l = \mathbf{E}\left( \{f_i\}_{i=1}^l ;\, \theta_{\mathbf{E}} \right),
\end{equation}
where $\theta_{\mathbf{E}}$ are trainable parameters of the transformer encoder. 

Although the global embedding sequence contains sufficient information to represent airfoils, directly concatenating them as the airfoil shape embedding is suboptimal. This representation resides in a high-dimensional space with inconsistent dimensions, increasing the difficulty of learning meaningful embeddings and impeding efficient training. To gather the global embedding sequence in a unified latent space, the mean pooling operation is applied to obtain the airfoil shape embedding:
\begin{equation}
    z = \frac{1}{l} \sum_{i=1}^{l} g_i.
\end{equation}

\subsubsection{Shape Decoder}
\label{sec:method shape decoder}

\begin{figure*}[htbp]
    \centering
    \includegraphics[width=1.\textwidth]{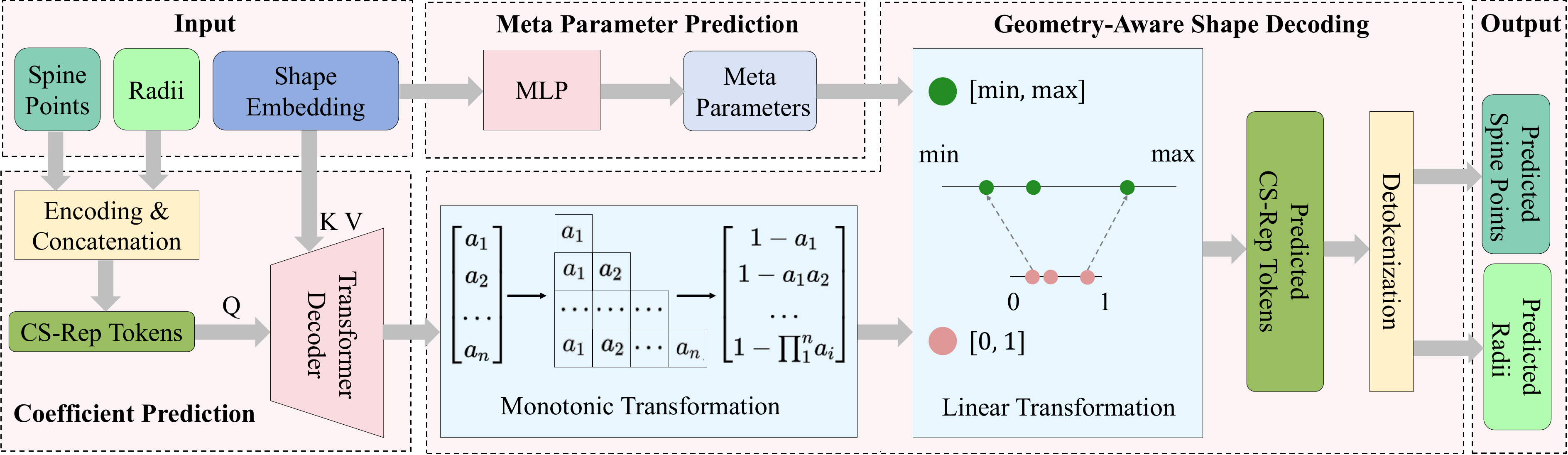}    
    \caption{The network for decoding airfoil shape embeddings into airfoils in the CS-Rep format.}
\label{fig:net decoder}
\end{figure*}

From an airfoil shape embedding, the shape decoder is designed to decode it into airfoils in the CS-Rep format, as shown in Fig.~\ref{fig:net decoder}. It consists of a meta sub-decoder and a coefficient sub-decoder: the meta sub-decoder predicts global information of an airfoil, denoted as meta parameters; based on these parameters, the coefficient sub-decoder generates the full set of components for CS-Rep, namely spine points and their corresponding radii.

\textbf{Meta sub-decoder}. The meta sub-decoder takes as input the airfoil shape embedding and predicts the meta parameters of an airfoil. Formally, these parameters are defined as the spine points and radii at the start, end, and extremum positions:
\begin{equation}
M = \left\{ \begin{array}{cccc}
    x_1 & y_1 & \Delta y_1 & r_1 \\
    x_{pos_p} & y_{pos_p} & \Delta y_{pos_p} & r_{pos_p} \\
    x_{pos_r} & y_{pos_r} & \Delta y_{pos_r} & r_{pos_r} \\
    x_n & y_n & \Delta y_{n-1} & r_n
\end{array} \right\}
\end{equation}
where $pos_p$ and $pos_r$ are the extremum positions of the spine the radius sequences, respectively.
Since these meta parameters are independent and do not exhibit causal dependencies, they can be predicted simultaneously with a simple MLP. Specifically, the MLP consists of two blocks, each containing a linear layer, a layer normalization, and an activation function, followed by a final linear projection that outputs the meta parameters.

\textbf{Coefficient sub-decoder}. Given the airfoil shape embedding and meta parameters, the coefficient sub-decoder will generate all spine points and their corresponding radii. To model the sequential dependencies within them, we adopt the transformer decoder in an autoregressive manner. Because the x-coordinate of the spine sequence can be directly generated using meta parameters by the uniformly spaced configuration, the token for the transformer decoder is defined as the y-coordinate of the spine and the corresponding radius:
\begin{equation}
    T = \{(y_i, r_i)\}_{i=1}^{n}.
\end{equation}
Note that the end and extremum positions of the token sequence should be determined before the decoding process. It can be achieved by using meta parameters:
\begin{equation}
\begin{aligned}
    n &= \frac{x_n - x_1}{\delta x} + 1 \\
    pos_p &= \frac{x_{pos_p} - x_1}{\delta x} + 1, \\
    pos_r &= \frac{x_{pos_r} - x_1}{\delta x} + 1
\end{aligned}
\end{equation}

Assume that the transformer decoder $\mathbf{D}$ has already generated $\tau$ tokens, denoted by $\hat{T}_{\tau}$, the goal is to model the next token. However, directly generating the next token fails to automatically guarantee the geometric characteristics in Sec.~\ref{sec:method representation}. In contrast, the transformer decoder generates two initial coefficients as described in Eq.~\eqref{eq:cumulative trick}:
\begin{equation}
\hat{\tilde{u}}_t,\hat{\tilde{v}}_t = \mathbf{D}\left( \hat{K}_{t-1}, \bar{g};\, \theta_{\mathbf{D}_{\text{coef}}} \right),
\end{equation}
Then, using the meta parameters, they can be reformulated by Eq.~\eqref{eq:piecewise reformulation} and ~\eqref{eq:cumulative trick} to obtain the next token:
\begin{equation}
\hat{u_i} = 1 - \prod_{j=1}^i \hat{\tilde{u}}_i, \,\,\, \hat{v}_i = 1 - \prod_{j=1}^i \hat{\tilde{v}}_i
\end{equation}
\begin{equation}
\begin{aligned}
\Delta \hat{y}_i &= 
\begin{cases}
\Delta y_1 + \hat{u}_i \, (\Delta y_{pos_p} - \Delta y_1), & \text{if } i \leq pos_p \\
\Delta y_{pos_p} + \hat{u}_i \, (\Delta y_{n-1} - \Delta y_{pos_p}), & \text{otherwise}
\end{cases} \\
\hat{r}_i &= 
\begin{cases}
r_1 + \hat{v}_i \, (r_{pos_r} - r_1), & \text{if } i \leq pos_r \\
r_{pos_r} + \hat{v}_i \, (r_n - r_{pos_r}), & \text{otherwise}
\end{cases}
\end{aligned}
\end{equation}
\begin{equation}
    \hat{y}_i = y_1 + \sum\limits_{j=1}^{i-1}\Delta y_j
\end{equation}

\textbf{Losses}. The two shape sub-decoders are trained jointly with the shape encoder. Specifically, in the training phase, the coefficient sub-decoder takes as input the ground truth of meta parameters, instead of the predictions of the meta sub-decoder. Because of the decoupled generation of the two sub-decoders, the reconstruction losses consist of two parts: the cross-entropy loss between the predictions and ground truth of quantized meta parameters by one-hot encoding for the meta sub-decoder, and the mean squared error between the predictions and ground truth of CS-Rep for the coefficient sub-decoder:
\begin{equation}
    L_{\text{recon}} = \lambda_1 \cdot \operatorname{CE}(M, \hat{M}) + \lambda_2 \cdot \operatorname{MSE}(T, \hat{T}),
\end{equation}
where $\lambda_1$ and $\lambda_2$ are the weights. In addition, to keep the predicted CS-Rep parameters within the feasible sweeping domain (note the square root in Eq.~\eqref{eq:radius direction}), an additional loss is introduced, as follows:
\begin{equation}
\label{eq:auxiliary loss}
    L_\text{auxi} = \lambda_3 \cdot \text{LeakyReLU}(({\Delta r_i})^2 - ({\Delta x_i})^2 - ({\Delta y_i})^2).
\end{equation}
The optimal performance is obtained with $\lambda_1 = 1 \times10^{-3}$, $\lambda_2 = 1$, and $\lambda_3 = 1 \times 10^{-6}$.

\begin{figure*}[ht]
    \centering
    \includegraphics[width=0.65\textwidth]{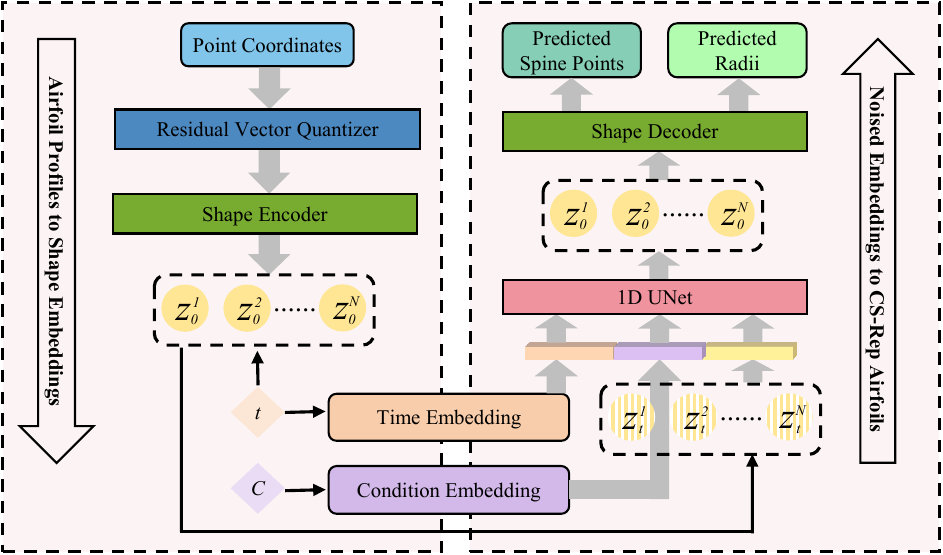}    
    \caption{The overall airfoil generation network architecture.}
\label{fig:net diffusion}
\end{figure*}

\subsection{Airfoil Shape Generation}
\label{sec:method generation}

To generate airfoils, the diffusion model is utilized to sample airfoil shape embeddings, which can be converted into airfoils in the CS-Rep format by the shape decoder (Fig.~\ref{fig:net diffusion}). In addition to unconditional generation, conditional mechanisms are introduced to establish an association between aerodynamic performance and airfoil shape embeddings, enabling performance-aware airfoil generation.

\subsubsection{Unconditional Generation}

For unconditional generation, we adopt the denoising diffusion probabilistic model (DDPM)~\cite{ho2020denoising}, a fundamental instance of diffusion models. DDPM learns the data distribution of airfoil shape embeddings through two processes: a forward process and a reverse process. The forward process gradually perturbs the data into noise, while the reverse process recovers corrupted data by learning an iterative denoising procedure.  

In the forward process, the airfoil shape embedding $z$ is perturbed by adding Gaussian noise step by step:
\begin{equation}
q(z_t \mid z_{t-1}) = \mathcal{N} \left( z_t; \sqrt{1 - \beta_t} \cdot z_{t-1}, \, \beta_t  \mathbf{I} \right)
\end{equation}
where $t$ denotes the time step, and $\beta_t$ denotes the weight of noise in $[0, 1]$. Then, the reverse process learns to invert the forward process by progressively denoising the fully noised embedding back into its original form. Specifically, the denoising step also follows a Gaussian distribution to remove noise gradually. Given the partially noised embedding $z_t$ and the time step $t$, a 1D UNet is utilized to predict a small noise to remove, which can derive the mean and variance of the Gaussian distribution, as follows (refer to ~\cite{ho2020denoising} for details of the derivation):
\begin{equation}
p_{\theta}(z_{t-1} \mid z_t) = \mathcal{N}\left(z_{t-1}; \frac{1}{\sqrt{\alpha_t}}\left(z_t - \frac{\beta_t}{\sqrt{1-\bar{\alpha}_t}}\epsilon_{\theta_{DDPM}}(z_t, t)\right), \beta_t\mathbf{I}\right) .
\end{equation}
where
\begin{equation}
\alpha_t = 1 - \beta_t, \,\,\, \bar{\alpha}_i = \prod_{j=1}^{t}\alpha_j.
\end{equation}

To train the DDPM, the mean squared distance between the predicted denoised embedding and the ground truth obtained from the forward process is the loss of each prediction. It can be simplified as the mean squared distance between the predicted noise and pure noise sampled from the standard normal distribution (refer to ~\cite{ho2020denoising} for details). In the inference stage, the DDPM samples pure noise from the standard normal distribution and passes it through the denoising steps, generating new airfoil shape embeddings that statistically align with the distribution of training data.

\subsubsection{Conditional Generation}

While the standard DDPM can generate airfoils, it lacks an explicit mechanism to control aerodynamic performance. To enable controllable generation, we employ a conditional DDPM (CDDPM) with classifier-free guidance (CFG)~\cite{ho2021classifier} to steer the reverse diffusion process toward desired aerodynamic performance.

\textbf{CDDPM}. Unlike the standard DDPM, CDDPM incorporates conditional information to guide the generation process. In this work, the conditions are formulated as discrete classes, each corresponding to a specific range of aerodynamic performance, characterized by the lift and drag coefficients. To control the generation process, the performance class $C$ is incorporated into the input of the denoising neural network to predict the noise residual:
\begin{equation*}
    \epsilon_{\theta_{CDDPM}}(z_t, t, C). 
\end{equation*}

\textbf{CFG}. Instead of directly using the trained denoising neural network in CDDPM, the CFG technique designs a new noise predictor to enhance controllability. Specifically, it leverages a linear combination of noise predictions from both the unconditional and conditional DDPMs. This guides the denoising trajectory toward the region of the data distribution that is strictly aligned with the prescribed performance class. Formally, the new noise predictor by CFG is represented as:
\begin{equation}
    \epsilon_{CFG}(z_t, t, C) = (1 + \omega) \epsilon_{\theta_{CDDPM}}(z_t, t, C) - \omega  \epsilon_{\theta_{DDPM}}(z_t, t).
\end{equation}
where $\omega$ is a hyperparameter that controls the strength of the guidance. An optimal value of $\omega = 3$ is observed in our experiments.

To mitigate the training costs associated with maintaining two separate noise predictors, a unified noise predictor is designed for both the DDPM and CDDPM. Building upon the CDDPM, an additional null class (represented by a zero vector, $\mathbf{0}$) is introduced to represent the marginal data distribution. Accordingly, the noise prediction for CFG is reformulated as:
\begin{equation}
    \epsilon_{CFG}(z_t, t, C) = (1 + \omega) \epsilon_{\theta_{CDDPM}}(z_t, t, C) - \omega  \epsilon_{\theta_{CDDPM}}(z_t, t, \mathbf{0}).
\end{equation}

\section{Results and Discussion}
\label{sec:result}

In this section, experiments are conducted to evaluate the proposed AirfoilGen method. The experimental setup is detailed in Sec.~\ref{sec:exp-setup}, followed by the ablation study in Sec.~\ref{sec:exp-ablation}. Results of reconstruction, unconditional generation, and conditional generation are presented in Sec.~\ref{sec:exp-reconstruction}, Sec.~\ref{sec:exp-unconditional generation}, and Sec.~\ref{sec:exp-conditional generation}, respectively. Finally, comparisons with existing airfoil generation methods are discussed in Sec.~\ref{sec:exp-comp}.

\subsection{Setup}
\label{sec:exp-setup}

\textbf{Dataset}. A dataset of 25,057 airfoil profiles with their CS-Rep and aerodynamic performance has been built. Specifically, we uniformly sample the parametrized curves of the NACA-4 and NACA-5 series by arc length to obtain airfoil profiles, and adopt the Euclidean distance transform to obtain their CS-Rep, namely, spine points and radii. Aerodynamic performance, including lift and drag coefficients, is evaluated by NeuralFoil~\cite{sharpe2025neuralfoil} in the following operating conditions: Reynolds number $Re = 2 \times 10^6$, Mach number $Ma = 0$, and angle of attack $\alpha = 0^\circ$. To further support the training of the diffusion model,  we augment the dataset with synthetically generated samples, expanding its size to more than 200,000 instances. 

\textbf{Training}. AirfoilGen is implemented in PyTorch and trained on an NVIDIA GeForce RTX 4090D GPU. Specifically, the residual vector quantization module is trained by the Adam optimizer with an initial learning rate of $1 \times 10^{-4}$ and a batch size of 1024 in 200 epochs. The airfoil shape autoencoder is trained by the AdamW optimizer with an initial learning rate of $3 \times 10^{-5}$ and a batch size of 128 in 1500 epochs. The diffusion model, with a linear scheduler of beta range $[0.0001, 0.02]$, is trained by the AdamW optimizer with an initial learning rate of $8 \times 10^{-4}$ and a batch size of 1024 in 2000 epochs.

\textbf{Evaluation metrics}. For reconstruction, two metrics are used to evaluate the similarity between reconstructed airfoils and ground truths, as follows:
\begin{itemize}
    \item \textbf{Chamfer Distance (ChD)} measures the average deviations between a reconstructed airfoil $\hat{P}$ and its corresponding ground truth $P$, defined as:
    \begin{equation*}
       \frac{1}{2}\left(\frac{1}{|P|} \sum_{p \in P} \min_{\hat{p} \in \hat{P}} \|p - \hat{p}\|_2 + \frac{1}{|\hat{P}|} \sum_{\hat{p} \in \hat{P}} \min_{p \in P} \|\hat{p} - p\|_2 \right).
    \end{equation*}

    \item \textbf{Hausdorff Distance (HD)} measures the maximum deviations between a reconstructed airfoil $\hat{P}$ and its corresponding ground truth $P$, defined as:
    \begin{equation*}
       \max \left\{ \max_{p \in P} \min_{\hat{p} \in \hat{P}} \|p - \hat{p}\|_2, \max_{\hat{p} \in \hat{P}} \min_{p \in P} \|\hat{p} - p\|_2 \right\}.
    \end{equation*}
\end{itemize}
For generation, another set of metrics is used to analyze the distribution of generated airfoils, as follows:
\begin{itemize}
    \item \textbf{Fidelity} refers to the average Hausdorff distance between each generated sample and its closest neighbor in the dataset.
    \item \textbf{Diversity} refers to the average Hausdorff distance between every pair of generated samples.
\end{itemize}

\subsection{Ablation Study}
\label{sec:exp-ablation}

In this subsection, we first examine the impact of several network components on learning airfoil shape embeddings. The ablation study is structured as follows:
\begin{enumerate}[label=(\alph*)]
    \item \textbf{AirfoilGen:} The full autoencoder used in AirfoilGen, including all modules.
    \item \textbf{w/o RVQ:} Remove the residual vector quantization module (Sec.~\ref{sec:method rlfq}); use one-hot encoding instead.
    \item \textbf{w/o auxiliary loss:} Remove the auxiliary loss in Eq.~\eqref{eq:auxiliary loss} during training.
\end{enumerate}

The ablation results, summarized in Table~\ref{tab:ae_component_ablation}, demonstrate that the full AirfoilGen autoencoder consistently yields the best performance. This observation underlines the effectiveness of the proposed autoencoder network detailed in Sec.~\ref{sec:method embedding}. 

\begin{table}[htbp]
\caption{Results of ablation studies on learning airfoil shape embeddings.} \vspace{4pt}
\centering
\setlength\extrarowheight{2pt}
\begin{tabularx}{0.5\textwidth}{
    >{\raggedright\arraybackslash}>{\hsize=1.\hsize\linewidth=\hsize}X
    >{\raggedright\arraybackslash}>{\hsize=.6\hsize\linewidth=\hsize}X
    >{\raggedright\arraybackslash}>{\hsize=.6\hsize\linewidth=\hsize}X
   }
    \Xhline{1pt}
    Model & \makecell[l]{$\text{ChD}_\textit{Profile}$} & \makecell[l]{$\text{HD}_\textit{Profile}$} \\
    \Xhline{0.5pt}
    AirfoilGen  & \textbf{4.11e-3} & \textbf{1.71e-3} \\
    \textbf{w/o} RVQ & 4.23e-3 & 1.76e-3 \\
    \textbf{w/o} auxiliary loss & 4.17e-3 & 1.72e-3 \\    
    \Xhline{1pt}
\end{tabularx}
\label{tab:ae_component_ablation}
\end{table}

\begin{figure*}[!h]
    \centering
    \includegraphics[width=0.95\textwidth]{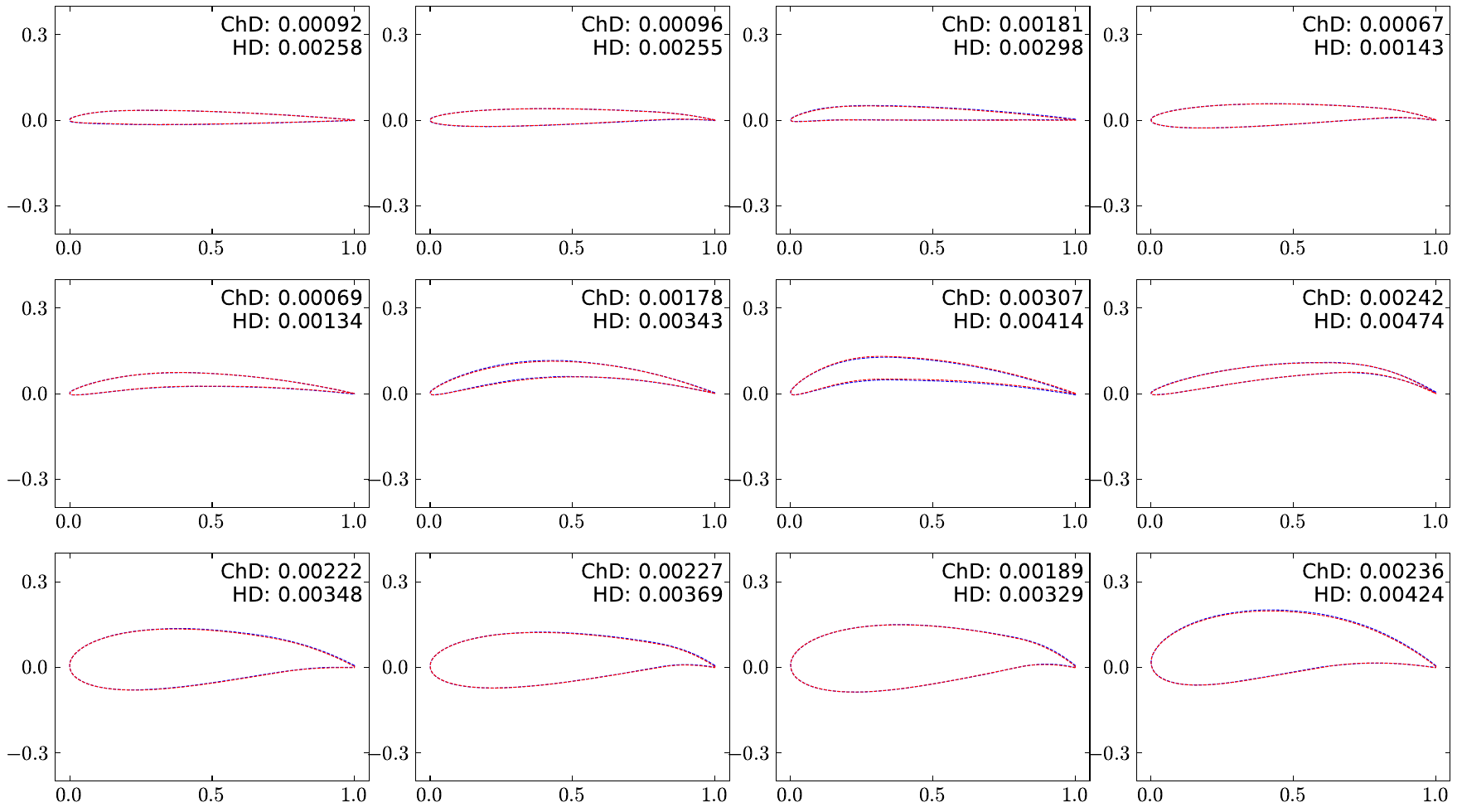}    
    \caption{
    Reconstruction examples of diverse airfoil shapes. The blue and red curves indicate the reconstructed airfoils and the ground truths, respectively. 
    }
\label{fig:representative reconstruction examples}
\end{figure*}

\begin{figure*}[!h]
    \centering
    \includegraphics[width=0.85\textwidth]{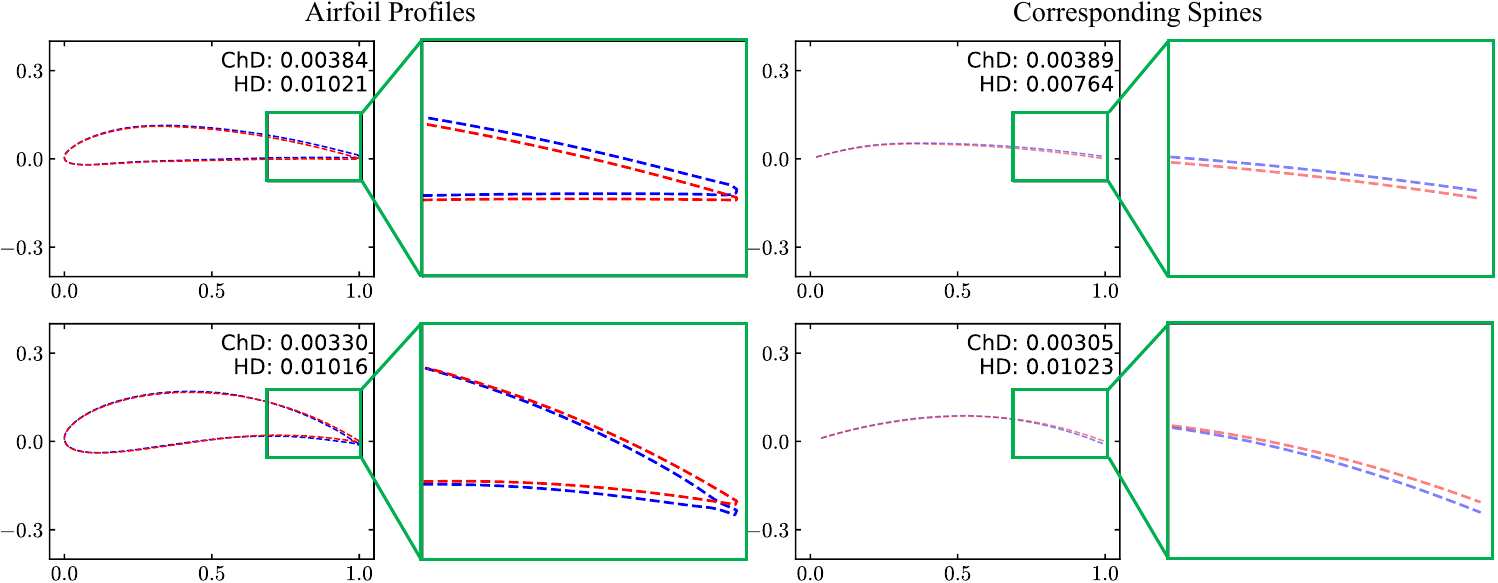}    
    \caption{Typical examples of large reconstruction deviations. The blue and red curves indicate the reconstructed airfoils and the ground truths, respectively.
    }
\label{fig:deviation reconstruction example}
\end{figure*}

Furthermore, the role of the autoencoder is assessed in airfoil generation. Specifically, two diffusion models are compared: one trained on the learned airfoil embeddings and the other trained without them. As shown in Table~\ref{tab:ae_generation_ablation}, removing the autoencoder reduces the success rate for satisfying the target aerodynamic conditions. This result suggests that the learned embeddings provide a more structured space for diffusion models to capture the airfoil shape distribution and align generated samples with the target performance.

\begin{table}[htbp]
\caption{Results of ablation studies on generating airfoils.} \vspace{4pt}
\centering
\setlength\extrarowheight{2pt}
\begin{tabularx}{0.5\textwidth}{
    >{\raggedright\arraybackslash}>{\hsize=1.\hsize\linewidth=\hsize}X
    >{\raggedright\arraybackslash}>{\hsize=.6\hsize\linewidth=\hsize}X
    >{\raggedright\arraybackslash}>{\hsize=.6\hsize\linewidth=\hsize}X
   }
    \Xhline{1pt}
    Model & \makecell[l]{$\text{Acc}_\textit{mean}$ ($\%$)} & \makecell[l]{$\text{Acc}_\textit{worst}$ ($\%$)} \\
    \Xhline{0.5pt}
    AirfoilGen  & \textbf{98.41} & \textbf{93.07} \\
    \textbf{w/o} autoencoder & 13.51 & 0 \\
    \Xhline{1pt}
\end{tabularx}
\label{tab:ae_generation_ablation}
\end{table}

\subsection{Reconstruction}
\label{sec:exp-reconstruction}

The reconstruction task assesses the autoencoder's encoding capability by comparing original airfoils with their counterparts decoded from airfoil shape embeddings. Table~\ref{tab:quantitative reconstruction results} presents the quantitative reconstruction results, showing that the autoencoder achieves consistently low errors across the training, validation, and test splits.

{\setlength{\intextsep}{0pt}
\begin{table}[htbp]
\caption{Evaluations of reconstruction errors across training, validation, and test splits.} \vspace{2pt}
\centering
\setlength\extrarowheight{2pt}
\begin{tabularx}{0.5\textwidth}{
    >{\raggedright\arraybackslash}>{\hsize=.2\hsize\linewidth=\hsize}X
    >{\raggedright\arraybackslash}>{\hsize=.4\hsize\linewidth=\hsize}X
    >{\raggedright\arraybackslash}>{\hsize=.4\hsize\linewidth=\hsize}X
    >{\raggedright\arraybackslash}>{\hsize=.4\hsize\linewidth=\hsize}X
    >{\raggedright\arraybackslash}>{\hsize=.4\hsize\linewidth=\hsize}X
   }
    \Xhline{1pt}
    Split & \makecell[l]{$\text{ChD}_\textit{CS-Rep}$} & \makecell[l]{$\text{HD}_\textit{CS-Rep}$} & \makecell[l]{$\text{ChD}_\textit{Profile}$} & \makecell[l]{$\text{HD}_\textit{Profile}$} \\
    \Xhline{0.5pt}
    Train & 1.10e-3 & 2.52e-3 & 1.59e-3 & 3.74e-3 \\
    Val   & 1.26e-3 & 2.91e-3 & 1.68e-3 & 4.02e-3 \\
    Test  & 1.29e-3 & 3.02e-3 & 1.71e-3 & 4.11e-3 \\
    \Xhline{1pt}
\end{tabularx}
\label{tab:quantitative reconstruction results}
\end{table}
}

Beyond quantitative evaluation, the reconstruction quality is qualitatively assessed through visual inspection. Fig.~\ref{fig:representative reconstruction examples} presents reconstruction examples across diverse shapes, all of which accurately capture the original shapes. In contrast, certain cases exhibit noticeable deviations from the target shapes, particularly in the trailing edge regions, as shown in Fig.~\ref{fig:deviation reconstruction example}. These deviations are likely due to error accumulation in spine points during the sequential generation process.

\subsection{Unconditional Generation}
\label{sec:exp-unconditional generation}

For unconditional generation, 8192 noise vectors are first sampled from a standard Gaussian distribution. Then, these vectors are fed into the diffusion model to generate airfoil shape embeddings. Finally, these embeddings are translated  into airfoils in the CS-Rep format by the decoder.

\begin{figure*}[htb]
    \centering
    \includegraphics[width=0.85\textwidth]{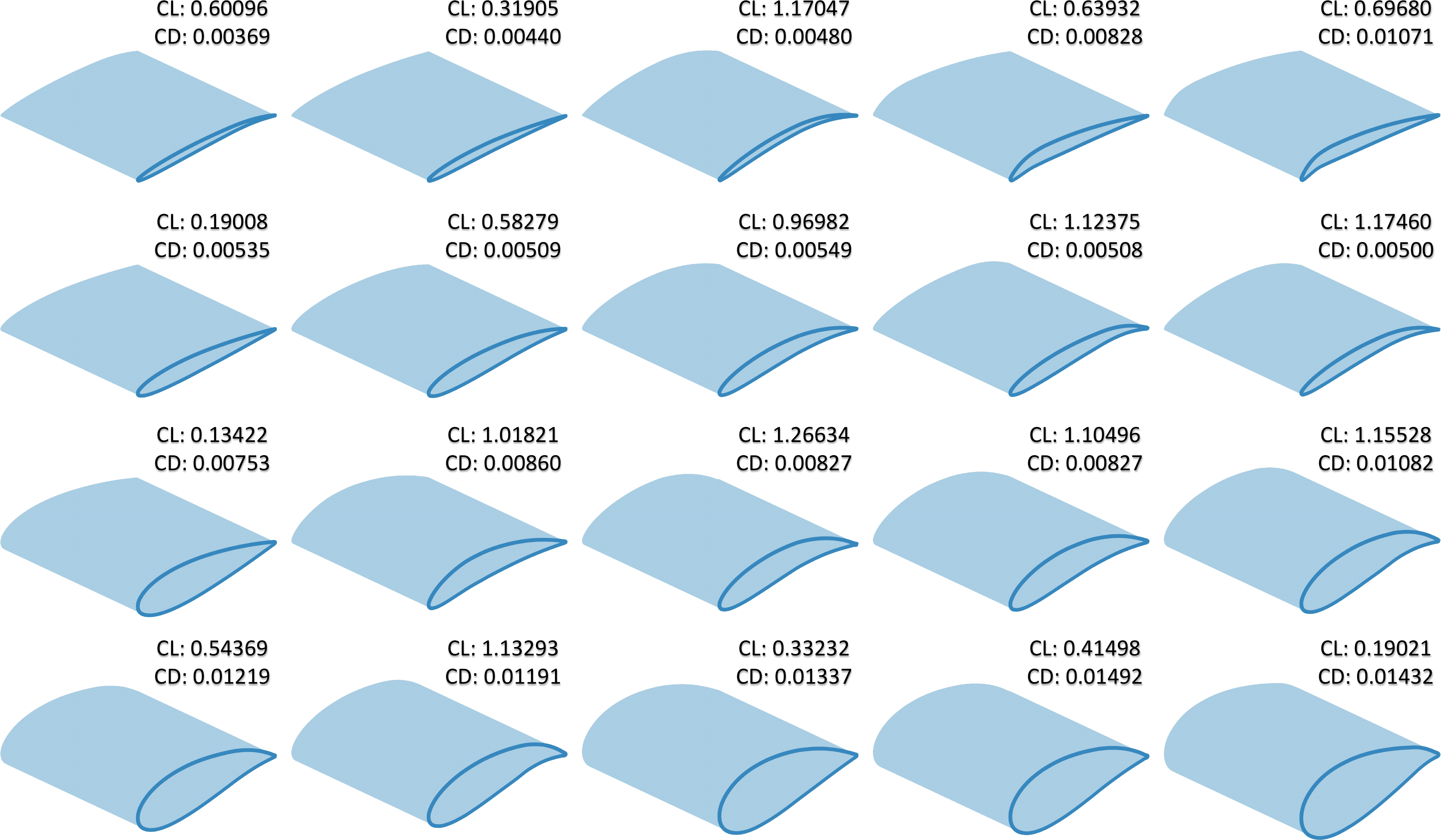}    
    \caption{Unconditional generation results of AirfoilGen. CL and CD denote the lift and drag coefficients evaluated by NeuralFFoil~\cite{sharpe2025neuralfoil}.}
\label{fig:unconditional examples}
\end{figure*}

Fig.~\ref{fig:unconditional examples} shows the representative airfoils generated in the unconditional setting. The fact that all generated samples maintain essential airfoil characteristics supports the effectiveness of the proposed CS-Rep in restricting generated airfoils to the prescribed valid shape space. Furthermore, the diverse range of airfoil shapes produced suggests that the diffusion model effectively captures the underlying data distribution rather than merely memorizing a limited set of training data.

\subsection{Conditional Generation}
\label{sec:exp-conditional generation}

For conditional generation, the aerodynamic performance metrics, namely lift and drag coefficients, are each uniformly divided into five numerical ranges, resulting in a total of 25 performance classes. For each class, 1024 airfoils are generated by conditioning the diffusion process on the respective classes. 

\begin{figure*}[htb]
    \centering
    \includegraphics[width=0.85\textwidth]{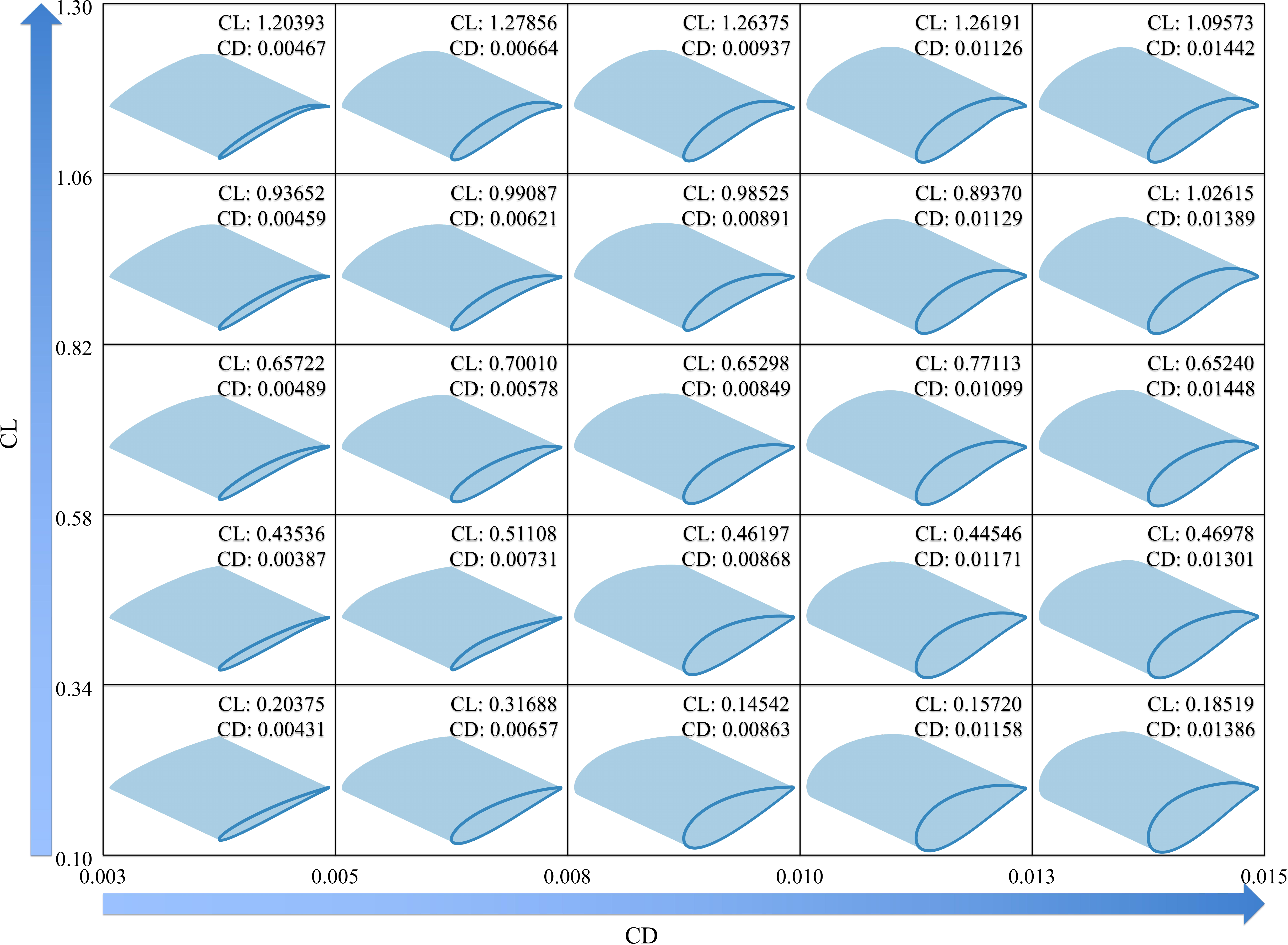}    
    \caption{ Conditional generation results of AirfoilGen for all performance classes. CL and CD denote the lift and drag coefficients evaluated by NeuralFoil~\cite{sharpe2025neuralfoil}.}
\label{fig:conditional examples}
\end{figure*}

Fig.~\ref{fig:conditional examples} presents typical examples generated under each performance class. Across all classes, the generated airfoils conform to predetermined performance requirements, demonstrating AirfoilGen's ability to produce performance-aware airfoils.

\begin{figure*}[htbp]
    \centering
    \includegraphics[width=0.8\textwidth]{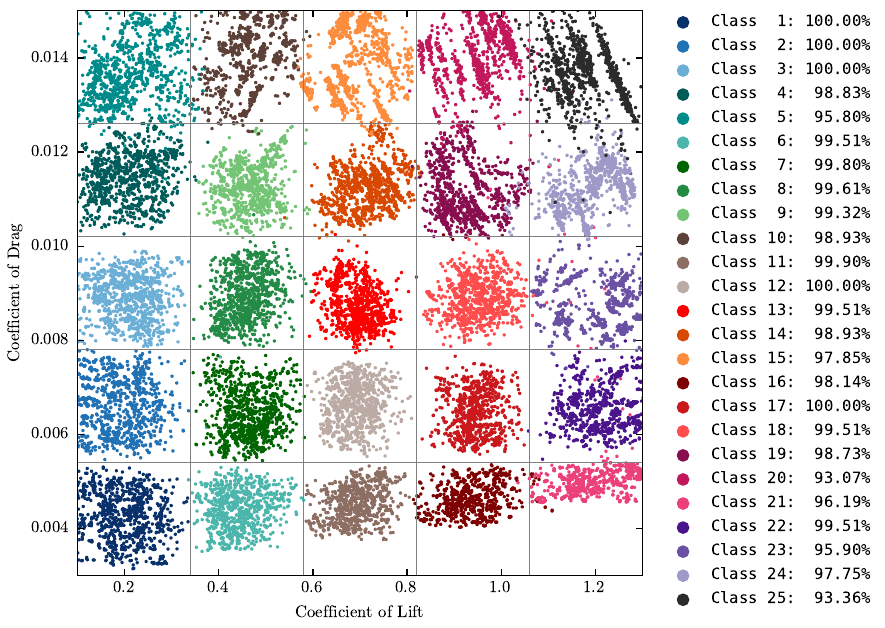}    
    \caption{Aerodynamic performance distribution of airfoils generated by AirfoilGen. Each point represents one generated airfoil positioned according to its lift and drag coefficients.}
\label{fig:conditional cl-cd distribution}
\end{figure*}

To further evaluate AirfoilGen's controllability quantitatively, the consistency of the generated aerodynamic performance with the target performance classes is examined. Fig.~\ref{fig:conditional cl-cd distribution} demonstrates the joint distribution of lift and drag coefficients for all generated samples, indicating that the generated airfoils closely align with the target performance regions. 

\subsection{Comparisons}
\label{sec:exp-comp}

To further evaluate the effectiveness of the proposed method, AirfoilGen is compared with existing airfoil generation methods, focusing on both geometric shapes and aerodynamic performance. Table~\ref{tab:overall qualitative comparison} provides a qualitative comparison between AirfoilGen and these methods (i.e., BézierGAN~\cite{chen2020airfoil}, AirfoilDiffusion~\cite{graves2024airfoil}, and CEBGAN~\cite{chen2022inverse}). The results demonstrate that AirfoilGen is the only approach capable of achieving performance-aware generation while inherently guaranteeing geometric validity.

\subsubsection{Shape Comparison}

For shape comparison, AirfoilGen is compared with BézierGAN, AirfoilDiffusion, and CEBGAN. AirfoilGen, BézierGAN, and AirfoilDiffusion are trained on our augmented dataset, while CEBGAN is trained on its original dataset due to its specialized problem definition.

Table~\ref{tab:quantitative shape comparison} shows the quantitative results based on two complementary shape evaluation metrics, Fidelity and Diversity \footnote{CEBGAN is not included in the table because it is trained on its own dataset.}. Fidelity quantifies the proximity of generated airfoils to the training distribution, while Diversity measures the variety among generated samples. A lower Fidelity score indicates greater realism, and a higher Diversity score denotes broader coverage of the design space. AirfoilGen achieves superior scores across both metrics, highlighting its ability to capture the real design manifold while maintaining high sample variety.

\begin{table*}[htbp]
\caption{Comparisons of existing airfoil generation methods with AirfoilGen.} \vspace{4pt}
\centering
\setlength\extrarowheight{2pt}
\begin{tabularx}{1\textwidth}{
    >{\centering\arraybackslash}>{\hsize=.6\hsize\linewidth=\hsize}X
    >{\centering\arraybackslash}>{\hsize=.5\hsize\linewidth=\hsize}X
    >{\centering\arraybackslash}>{\hsize=.6\hsize\linewidth=\hsize}X
    >{\centering\arraybackslash}>{\hsize=.4\hsize\linewidth=\hsize}X
    >{\centering\arraybackslash}>{\hsize=.5\hsize\linewidth=\hsize}X
   }
    \Xhline{1pt}
    Model &\makecell[c]{No Self-intersection} &\makecell[c]{Smoothness Constraint} &\makecell[c]{No Distortion} &\makecell[c]{Performance Control}\\
    \Xhline{0.5pt}
    AirfoilGen &Yes&Yes&Yes&Yes\\
    BézierGAN~\cite{chen2020airfoil} &No&Yes&Yes&No\\
    CEBGAN~\cite{chen2022inverse} &Yes&Yes&No&Yes\\
    AirfoilDiffusion~\cite{graves2024airfoil} &No&No&Yes&Yes\\
    \Xhline{1pt}
\end{tabularx}
\label{tab:overall qualitative comparison}
\end{table*}

\begin{figure*}[htbp]
    \centering
    \includegraphics[width=0.8\textwidth]{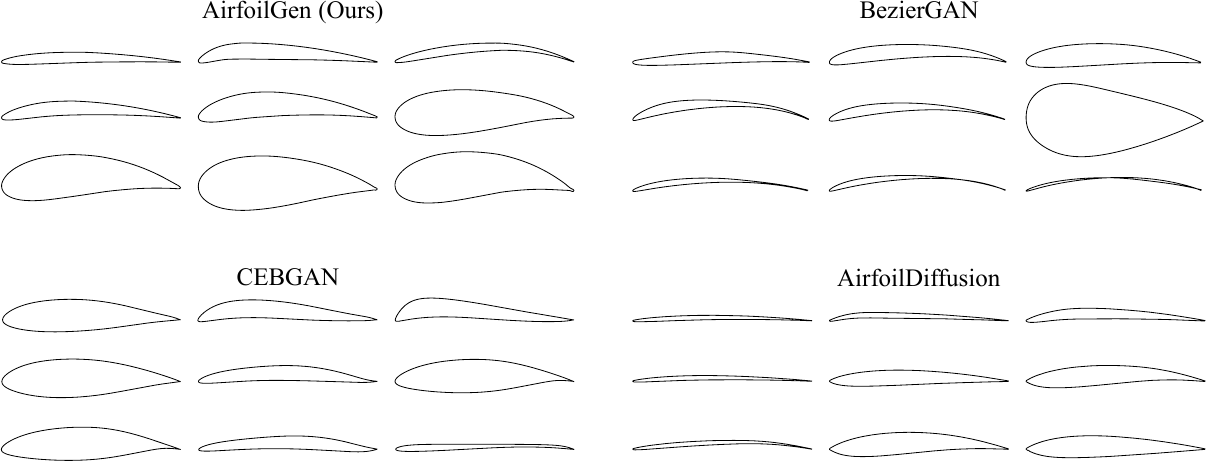} 
    \caption{Comparisons of generated airfoil shapes between AirfoilGen and existing methods.}
\label{fig:qualitative shape comparison}
\end{figure*}

Qualitative comparisons in Fig.~\ref{fig:qualitative shape comparison} demonstrate the geometric limitations of the compared methods. Specifically, BézierGAN results in self-intersections (the right-bottom case), CEBGAN exhibits structural distortions (the middle-bottom case), and AirfoilDiffusion suffers from surface overlapping at the trailing edge (the left-bottom case). These artifacts reflect the difficulty of maintaining geometric validity when generation is performed in an unconstrained or weakly constrained shape space. In contrast, AirfoilGen avoids such failures by generating airfoils in CS-Rep format, where the prescribed airfoil characteristics are enforced by construction.

\begin{table}[htbp]
\caption{Comparisons of shape distribution metrics between AirfoilGen and existing methods.} 
\vspace{4pt}
\centering
\setlength\extrarowheight{2pt}
\begin{tabularx}{0.5\textwidth}{
    >{\raggedright\arraybackslash}>{\hsize=1\hsize\linewidth=\hsize}X
    >{\raggedright\arraybackslash}>{\hsize=.4\hsize\linewidth=\hsize}X
    >{\raggedright\arraybackslash}>{\hsize=.4\hsize\linewidth=\hsize}X
 }
    \Xhline{1pt}
    Model & \makecell[l]{Fidelity $\downarrow$} & \makecell[l]{Diversity $\uparrow$} \\
    \Xhline{0.5pt}
    AirfoilGen & \textbf{5.85e-4} & \textbf{7.83e-2} \\
    BézierGAN & 1.82e-2 & 7.24e-2 \\
    AirfoilDiffusion & 9.28e-3 & 4.02e-2 \\
    \Xhline{1pt}
\end{tabularx}
\label{tab:quantitative shape comparison}
\end{table}

\begin{table*}[htbp]
\caption{Comparisons of generation accuracy ($\%$) across five categories between AirfoilGen, AirfoilDiff-CL, and AirfoilDiff-CD.} 
\vspace{4pt}
\centering
\setlength\extrarowheight{2pt}
\begin{tabularx}{0.75\textwidth}{
    >{\raggedright\arraybackslash}>{\hsize=1.1\hsize\linewidth=\hsize}X
    >{\centering\arraybackslash}>{\hsize=.55\hsize\linewidth=\hsize}X
    >{\centering\arraybackslash}>{\hsize=.55\hsize\linewidth=\hsize}X
    >{\centering\arraybackslash}>{\hsize=.55\hsize\linewidth=\hsize}X
    >{\centering\arraybackslash}>{\hsize=.55\hsize\linewidth=\hsize}X
    >{\centering\arraybackslash}>{\hsize=.55\hsize\linewidth=\hsize}X
}
    \Xhline{1pt}
    Model & \#0 & \#1 & \#2 & \#3 & \#4 \\
    \Xhline{0.5pt}
    AirfoilGen      & \textbf{100.00} & \textbf{99.80} & \textbf{99.51} & \textbf{98.73} & \textbf{93.36} \\
    AirfoilDiff-CL  & \textbf{100.00} & 35.35         & 0          & 0          & 0          \\
    AirfoilDiff-CD  & 99.90          & 0          & 0          & 0          & 0          \\
    \Xhline{1pt}
\end{tabularx}
\label{tab:generation_accuracy_comparison}
\end{table*}

\begin{table*}[htbp]
\caption{Comparison of shape optimization results initialized with airfoils generated by AirfoilGen and random initialization under the same target error tolerance.}
\vspace{4pt}
\centering
\setlength\extrarowheight{2pt}
\begin{tabularx}{0.95\textwidth}{
    >{\raggedright\arraybackslash}>{\hsize=1.2\hsize\linewidth=\hsize}X
    >{\centering\arraybackslash}>{\hsize=.9\hsize\linewidth=\hsize}X
    >{\centering\arraybackslash}>{\hsize=.55\hsize\linewidth=\hsize}X
    >{\centering\arraybackslash}>{\hsize=.55\hsize\linewidth=\hsize}X
    >{\centering\arraybackslash}>{\hsize=.7\hsize\linewidth=\hsize}X
    >{\centering\arraybackslash}>{\hsize=.7\hsize\linewidth=\hsize}X
}
    \Xhline{1pt}
    Initialization & Success Rate ($\%$) & Iterations & Time (s) & Target MAE$_{CL}$ & Target MAE$_{CD}$ \\
    \Xhline{0.5pt}
    AirfoilGen
    & \textbf{99.80}
    & \textbf{21} 
    & \textbf{38.62} 
    & \multirow{2}{*}{2e-3} 
    & \multirow{2}{*}{2e-3} \\
    
    Random Initialization 
    & 47.04
    & 47          
    & 91.27          
    &  
    &  \\
    \Xhline{1pt}
\end{tabularx}
\label{tab:shape_optimization_comparison}
\end{table*}

\begin{figure*}[!h]
    \centering
    \includegraphics[width=0.95\textwidth]{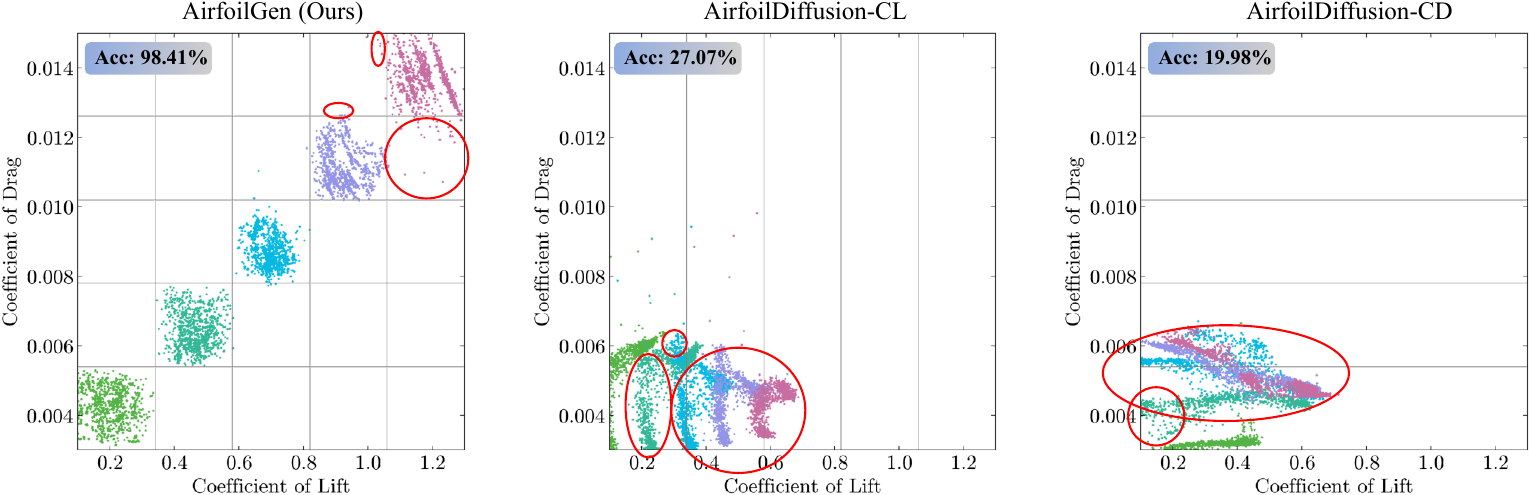}    
    \caption{ Comparisons of aerodynamic performance distributions between AirfoilGen and existing methods. Circled regions indicate samples that violate the predefined aerodynamic constraints.}
\label{fig:conditional performance distribution comparison}
\end{figure*}

\begin{figure*}[!h]
    \centering
    \includegraphics[width=0.75\textwidth]{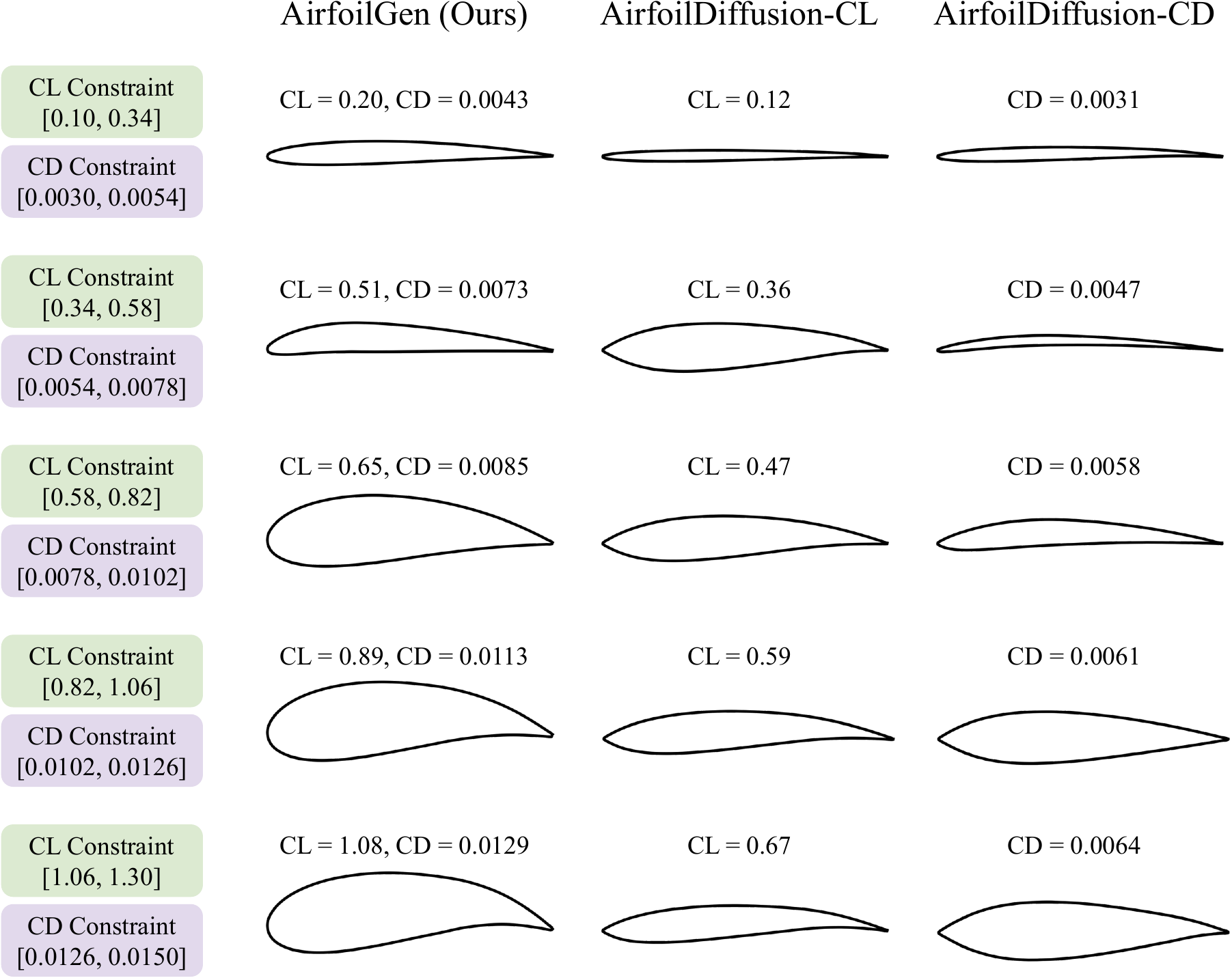}    
    \caption{Comparison of aerodynamic performance for airfoils generated by AirfoilGen and existing methods.}
\label{fig:conditional example comparison}
\end{figure*}

\subsubsection{Performance Comparison}

For performance comparison, AirfoilGen is compared with AirfoilDiffusion in performance control. Because AirfoilDiffusion supports only a single conditioning variable, two separate models are trained for the lift and drag coefficients, denoted as AirfoilDiffusion-CL and AirfoilDiffusion-CD. All models are trained on our augmented dataset.

Fig.~\ref{fig:conditional performance distribution comparison} compares the aerodynamic performance distributions of airfoils generated by AirfoilGen, AirfoilDiffusion-CL, and AirfoilDiffusion-CD across 5 target performance regions, while the generation accuracy of each performance class is summarized in Table~\ref{tab:generation_accuracy_comparison}. It is observed that AirfoilGen consistently aligns with the specified aerodynamic constraints for most samples. In contrast, AirfoilDiffusion-CL and AirfoilDiffusion-CD exhibit a narrower controllable range and frequently fail to reach the desired targets. These results are further illustrated in Fig.~\ref{fig:conditional example comparison}, where selected airfoils for each class are displayed alongside their corresponding target constraints (the leftmost column).

To further evaluate the practical value of AirfoilGen in obtaining airfoils with more specific aerodynamic performance targets, we conduct a shape optimization experiment using the generated airfoils as initial shapes. Specifically, we randomly select 10 target performance values from each of the 25 performance classes and, for each target value, generate 10 initial airfoil shapes with AirfoilGen for subsequent optimization. The results are compared with those obtained from randomly initialized airfoils, as reported in Table~\ref{tab:shape_optimization_comparison}. Compared with random initialization, AirfoilGen achieves a substantially higher success rate in reaching the same target error tolerance. This suggests that the generated airfoils provide more favorable initial shapes for aerodynamic optimization, making the subsequent optimization process much more reliable. The small fraction of failure cases may be attributed to the occasional instability of the numerical optimizer. In addition, AirfoilGen-based initialization requires fewer iterations and less optimization time than random initialization. Overall, these results demonstrate that AirfoilGen not only supports controllable airfoil generation, but also provides effective initialization for accurate and efficient downstream aerodynamic optimization.

\section{Conclusion}
\label{sec:conclusion}

In this paper, we propose the circle sweeping representation (CS-Rep), a scheme that ensures airfoil geometric validity through embedded constraints. Based on this representation, we develop AirfoilGen, a generative model for performance-aware airfoil design. Our approach utilizes an autoencoder to map airfoil geometries into a latent space, where latent vector embeddings are decoded back into the CS-Rep format to maintain geometric integrity automatically. Within this latent space, a conditional diffusion model is trained to guide the generation process toward target aerodynamic performance. To support the training of this model, a large-scale dataset containing over 200,000 airfoil designs and performance labels is also introduced. Experimental results demonstrate the performance of the proposed method in both unconditional and conditional generation tasks.

Despite its effectiveness, the proposed method has certain limitations. First, due to computational constraints, the current model uses 25 discrete performance classes, which limits the granularity of control over continuous aerodynamic values. Nevertheless, the shape optimization experiment shows that, even under this coarse conditioning setting, the generated airfoils can serve as high-quality initial designs and accelerate convergence toward more accurate target performance. This result suggests that a finer-grained conditioning scheme may further improve aerodynamic control precision. Future work could explore this direction while addressing the increased demands on data coverage, model capacity, and training cost.

Second, as illustrated in Fig.~\ref{fig:conditional cl-cd distribution}, for specific performance combinations (e.g., extreme CL/CD ratios), the generated airfoils may exhibit a performance shift from the target. This is likely due to the inherent long-tail distribution of the training dataset. We anticipate that the emergence of larger and more diverse airfoil datasets will further broaden the controllable design space of our method.

\section*{Acknowledgements}

This work has been funded by the ``Pioneer" and ``Leading Goose" R\&D Program of Zhejiang Province (No. 2024C01103), the National Natural Science Foundation of China (No. 62102355), and the Fundamental Research Funds for the Zhejiang Provincial Universities (No. K20250142, K20241957).

\section*{References}

\bibliographystyle{elsarticle-num}
\bibliography{mybibfile}

\end{document}